# How we can control the crack to propagate along the specified path feasibly?

Zhenxing Cheng, Hu Wang[*]

*State Key Laboratory of Advanced Design and Manufacturing for Vehicle Body, Hunan University, Changsha, 410082, P.R. Chin*

**Abstract** A controllable crack propagation (CCP) strategy is suggested. It is well known that crack always leads the failure by crossing the critical domain in engineering structure. Therefore, the CCP method is proposed to control the crack to propagate along the specified path, which is away from the critical domain. To complete this strategy, two optimization methods are engaged. Firstly, a back propagation neural network (BPNN) assisted particle swarm optimization (PSO) is suggested. In this method, to improve the efficiency of CCP, the BPNN is used to build the metamodel instead of the forward evaluation. Secondly, the popular PSO is used. Considering the optimization iteration is a time consuming process, an efficient reanalysis based extended finite element methods (X-FEM) is used to substitute the complete X-FEM solver to calculate the crack propagation path. Moreover, an adaptive subdomain partition strategy is suggested to improve the fitting accuracy between real crack and specified paths. Several typical numerical examples demonstrate that both optimization methods can carry out the CCP. The selection of them should be determined by the tradeoff between efficiency and accuracy.

## 1 Introduction

Generally, the internal crack propagation is a critical issue in the engineering practice due to its deep effect on the quality and stability of engineering structures. Therefore, predicting the path of crack propagation is significant for guaranteeing the safety or reliability of engineering structures. There are many numerical methods of simulating crack propagation. Such as finite element method (FEM) (Bouchard et al., 2003; Branco et al., 2015), extended finite element method (X-FEM) (Belytschko et al., 2009; Zeng et al., 2016), edge-based finite element method (ES-FEM) (G. R. Liu et al., 2011; Nguyen-Xuan et al., 2013), meshless method (Gu et al., 2011; Tanaka et al., 2015), and so on. The X-FEM might be the most popular method for crack propagation simulation due to its superiority of modeling both strong and weak discontinuities. Belytschko and Black proposed the initial idea of X-FEM at 1999 with minimal re-mesh (Belytschko et al., 1999). Then, Moës el al. (Belytschko et al., 2001) and Dolbow et al. (Dolbow et al., 1999) adopted the Heaviside function to enrichment function and 3D static crack was modeled by Sukumar et al. (Sukumar et al., 2000). Sequentially, the level set methods (LSMs) were applied to X-FEM which could easily track both the crack position and tips (Stolarska et al., 2001). Moreover, the X-FEM has much more applications (Ahmed et al., 2012; Areias et al., 2005; Belytschko et al., 2003; Chessa et al., 2002; Huynh et al., 2009; J.-H. Song et al., 2006; Sukumar et al., 2001; Zhuang et al., 2011; Zilian et al., 2008). More details of the development of X-FEM can be found in the literature (Abdelaziz et al., 2008; Belytschko et al., 2009; Fries et al., 2010).

It is well known that the internal crack propagation always leads the failure of engineering structure by crossing the critical domain of the structure. Therefore, if the crack doesn't cross the critical domain, the failure will not happen. Therefore, a controllable crack propagation method is proposed to control the crack propagation path and lead it propagate along the pre-defined path, so that the critical domain should not be crossed by the crack and the failure will not happen. In this study, the particle swarm optimization (PSO) method is used to obtain the suitable variables of design and the artificial neural network is used to improve the efficiency of PSO.

The PSO proposed by Kennedy and Eberhart is a popular metaheuristic algorithm which inspired by the social behavior of bird flocking (Kennedy et al., 1995). Later Kennedy and Eberhart suggested a developed version of PSO for discrete optimization (Kennedy et al., 1997). Shi and Eberhart improved the PSO by inertia weight (Shi et al., 1998). Recently, PSO has been applied to many fields, such as structural optimization (Vagelis et al., 2011), dynamic finite element model updating (Shabbir et al., 2015), vehicle engineering (Battaü et al., 2013), artificial neural network (Chatterjee et al., 2016; W. Sun et al., 2016) and so on (Amini et al., 2013; Amiri et al., 2012; Delice et al., 2014). Much more studies on PSO can be found in the literature (Eberhart et al., 2001; Ma et al., 2015; Poli et al., 2007; Tyagi et al., 2011).

Considering the optimization iteration is a time consuming process, an efficient reanalysis based X-FEM is used to calculate the crack propagation path, in which reanalysis methods are used to solve the equilibrium equations efficiently. Reanalysis (Kirsch, 2002), as a fast computational method, was suggested to predict the response of modified structures efficiently instead of full analysis. In recent decades, reanalysis methods have been well developed. Song et al. proposed a direct reanalysis algorithm to update the triangular factorization in sparse matrix solution (Q. Song et al., 2014). Liu et al. applied the Cholesky factorization to structural reanalysis (H. F. Liu et al., 2014). Huang and Wang solved the large-scale problems with local modification by the independent coefficient (IC) method (Huang et al., 2013). Zuo *et al.* applied the reanalysis method to the genetic algorithm (GA) (Zuo et al., 2011). Sun *et al*. extended the reanalysis method into a structural optimization process (R. Sun et al., 2014). To improve the efficiency of reanalysis method, He *et al.* developed a multiple-GPU based parallel IC reanalysis method (He et al., 2015). Based on these techniques, Wang *et al.* developed a CAD/CAE integrated parallel reanalysis design system (H. Wang et al., 2016).

In this study, a controllable crack propagation (CCP) method is proposed to control the crack propagation path and make the crack propagate along the specified path, so that the critical domain of engineering structure should not be destroyed by the crack. Moreover, considering the optimization iteration is a time consuming process, an efficient reanalysis based X-FEM is used to calculate the crack propagation path, in which the reanalysis solver is used to solve the equilibrium equations efficiently. Then the BPNN assisted PSO method should be used to obtain the optimal design variables to make the real crack path match with the specified path.

The rest of this paper is organized as follows. The basic theory of X-FEM is briefly introduced in Section 2. The details of the CCP method can be found in Section 3. Then, some numerical examples will be given to investigate the performance of the CCP method in Section 4. Finally, some conclusions are summarized in Section 5.

## 2 Basics theories of XFEM

### 2.1 X-FEM approximation

In the X-FEM, the displacement approximation consists of two parts: the standard finite element approximation and partition of unity enrichment. Define the displacement approximation of X-FEM as:

$$\mathbf{u}^h(\mathbf{x}) = \underbrace{\sum_{I \in \Omega} N_I(\mathbf{x})\mathbf{u}_I}_{\mathbf{u}^{standard}} + \underbrace{\sum_{J \in \Omega_E} \psi(\mathbf{x})N_J(\mathbf{x})\mathbf{q}_J}_{\mathbf{u}^{enrich}} \quad (1)$$

where $N_I$ and $\mathbf{u}_I$ denote the standard FEM shape function and nodal degrees of freedom (DOF) respectively. The $\psi(\mathbf{x})$ means enrichment function while the $\mathbf{q}_J$ is the additional nodal degree of freedom.

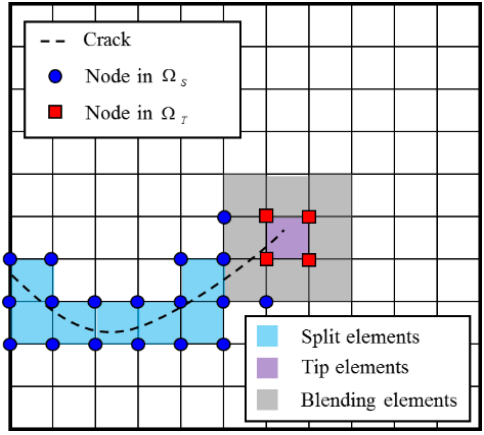

Fig. 1 An arbitrary crack line in a structured mesh

Consider an arbitrary crack in a structured mesh as shown in Fig. 1, and then Eq. (1) can be rewritten as

$$\mathbf{u}^h(\mathbf{x}) = \sum_{I \in \Omega} N_I(\mathbf{x})\mathbf{u}_I + \sum_{I \in \Omega_S} H_I(\mathbf{x}) N_I(\mathbf{x})\mathbf{a}_I + \sum_{I \in \Omega_T} \sum_{\alpha=1}^{4} \Phi_{I,\alpha}(\mathbf{x}) N_I(\mathbf{x})\mathbf{b}_I^\alpha, \quad (2)$$

where $\Omega$ is the solution domain, $\Omega_S$ is the domain cut by crack, $\Omega_T$ is the domain which crack tip located, $H(\mathbf{x})$ is the shifted Heaviside enrichment and $\Phi_\alpha(\mathbf{x})$ is the shifted crack tip enrichment. The details of $H(\mathbf{x})$ and $\Phi_\alpha(\mathbf{x})$ are given as following:

$$H(\mathbf{x}) = \begin{cases} +1 & \text{Above crack} \\ -1 & \text{Below crack} \end{cases}, \quad (3)$$

$$\{\Phi_\alpha(\mathbf{x})\}_{\alpha=1}^{4} = \sqrt{r}\left\{\sin\frac{\theta}{2}, \cos\frac{\theta}{2}, \sin\theta\sin\frac{\theta}{2}, \sin\theta\cos\frac{\theta}{2}\right\}. \quad (4)$$

The discrete X-FEM equations by substituting Eq.(2) into the principle of virtual work can be written as

$$\begin{bmatrix} \mathbf{K}_{uu} & \mathbf{K}_{ua} & \mathbf{K}_{ub} \\ \mathbf{K}_{ua}^T & \mathbf{K}_{aa} & \mathbf{K}_{ab} \\ \mathbf{K}_{ub}^T & \mathbf{K}_{ab}^T & \mathbf{K}_{bb} \end{bmatrix} \begin{Bmatrix} \mathbf{u} \\ a \\ \mathbf{b} \end{Bmatrix} = \begin{Bmatrix} \mathbf{F}_u \\ \mathbf{F}_a \\ \mathbf{F}_b \end{Bmatrix}, \quad (5)$$

where $\mathbf{K}_{uu}$ is the traditional finite element stiffness matrix, $\mathbf{K}_{ua}, \mathbf{K}_{aa}, \mathbf{K}_{ab}$ are components with Heaviside enrichment and $\mathbf{K}_{ub}, \mathbf{K}_{ab}, \mathbf{K}_{bb}$ are components with crack tip enrichment.

### 2.2 Crack propagation model

Generally, the direction and magnitude of crack propagation at each iteration are used to determine how the crack will propagate. The direction of crack propagation is found from the maximum circumferential stress criterion and the crack will propagate in the direction where $\sigma_{\theta\theta}$ is maximum (Erdogan et al., 1963). The angle of crack propagation is defined as

$$\theta = 2\arctan\frac{1}{4}\left(\frac{K_I}{K_{II}} - \text{sign } K_{II}\sqrt{\left(\frac{K_I}{K_{II}}\right)^2 + 8}\right), \quad (6)$$

where $\theta$ is defined in the crack tip coordinate system, $K_I$ and $K_{II}$ are the mixed-mode stress intensity factors. The details are given in the reference (Erdogan et al., 1963).

There are two main patterns when modeling crack growth. The first one assumes a constant increment of crack growth at each cycle (Dolbow et al., 1999) while the other option is to assume a constant number of cycles and apply a fatigue crack growth law to predict the crack growth increment for the fixed number of cycles (Gravouil et al., 2002). In this study, a fixed increment of crack growth $\Delta a$ is considered.

## 3 Controllable crack propagation method

### 3.1 Framework of the CCP method

As mentioned above, the CCP method is proposed to control the crack propagation path based on the X-FEM, reanalysis, and BPNN assisted PSO methods. The framework of CCP method is shown in Fig. 2.

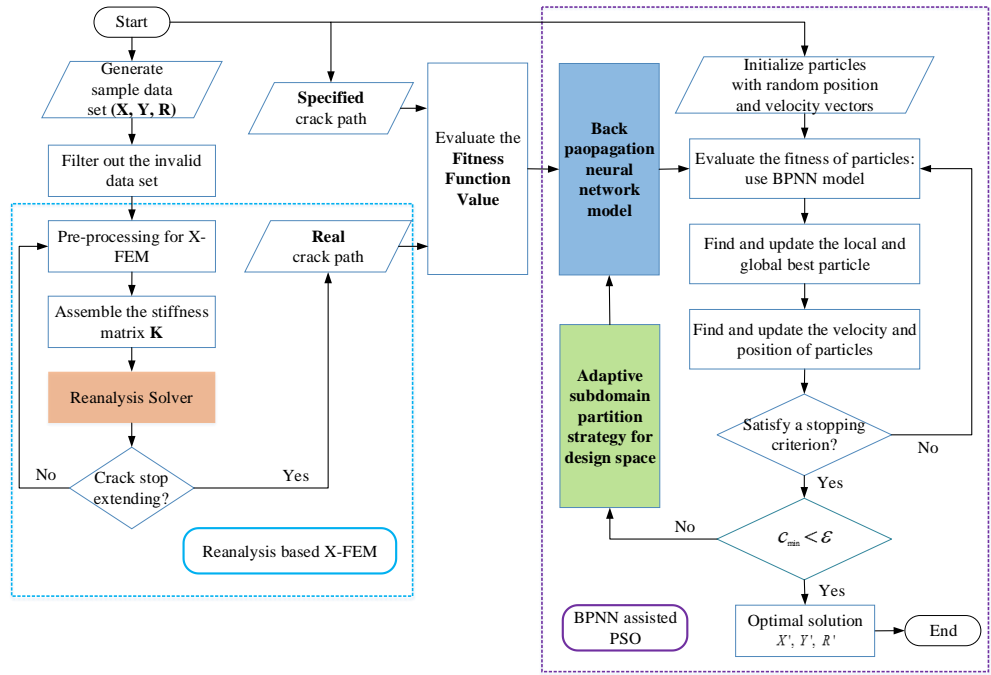

Fig. 2 The framework of the CCP method

It can be found that the CCP method mainly includes two parts: reanalysis based X-FEM and BPNN assisted PSO method. The first part, reanalysis based X-FEM is used to calculate the real crack path and the reanalysis solver is used to solve the equilibrium equations to improve the efficiency of the X-FEM. The second part, the BPNN assisted PSO is used to obtain the optimal design variables to make the real crack path match with the specified path. Moreover, the adaptive subdomain partition strategy is used to improve the fitting accuracy between real crack and specified paths. More details can be found in Section 3.2, 3.3 and 3.4.

**3.2 How to control the crack propagation path**

Assume an edge crack in a plate as shown in the left of Fig. 3, where the initial crack length is $a_0 = 10mm$, the force $F = 2 \times 10^4 N$, and linear elastic material behavior is assumed. The material is aluminum 7075-T6 with $E = 7.17 \times 10^4 MPa$, $v = 0.33$ and a plane strain state is considered. The increment of propagation $\Delta a = 1mm$. Calculate the crack propagation path by X-FEM without reanalysis solver (full analysis), and the crack path are shown in the right of Fig. 3.

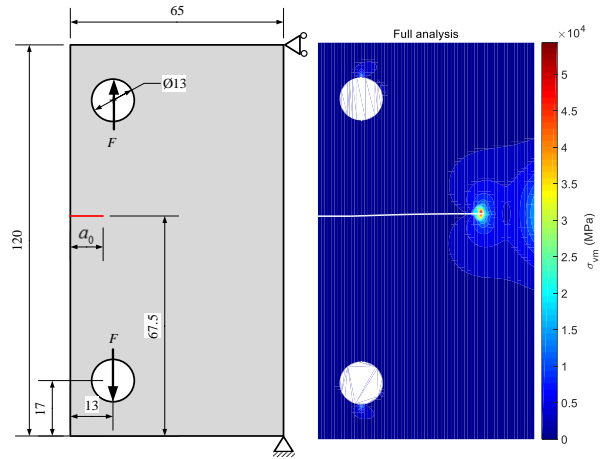

Fig. 3 An edge crack in a plate

Then if a hole was added in the plate as shown in the left of Fig. 4, the crack propagation path will change as the right of Fig. 4. Obviously, the crack propagation path can be influenced by the size, position and number of this hole. In order to clearly describe this property, several different results are shown in Fig. 5. It can be found that different arrangement of the holes obtained different crack propagation paths and the paths can be easily driven by arranging the holes.

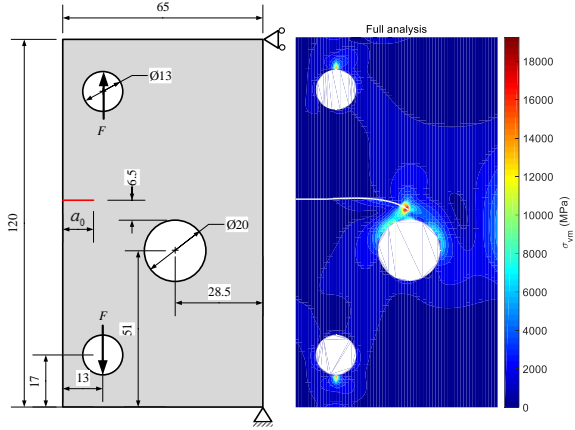

Fig. 4 An edge crack in a plate with a hole

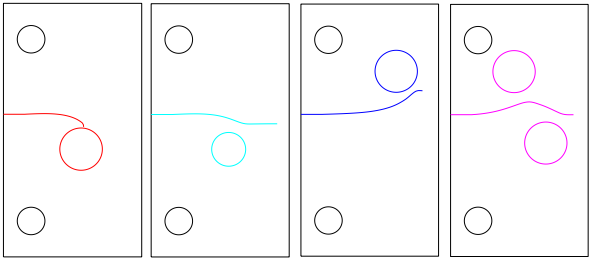

Fig. 5 The crack propagation paths of different arrangement of holes

## 3.3 The BPNN assisted PSO for crack path

Particle swarm optimization is a popular optimization algorithm, and a general algorithm of PSO is described as the following algorithm (Poli et al., 2007):

**Algorithm: General algorithm for PSO**

1: Initialize a population array of particles with random position and velocity vectors;
2: **Loop**
3: For each particle, evaluate the fitness by fitness function;
4: Compare particle's fitness evaluation with its $p_{best_i}$. If current value is better than $p_{best_i}$, then update the $p_{best_i}$ and $\vec{p}_i$;
5: Update the velocity and position of the particle according to the following equation:

$$\begin{cases} \vec{v}_i \leftarrow \vec{v} + \vec{U}(0,\phi_1) \otimes (\vec{p}_i - \vec{x}_i) + \vec{U}(0,\phi_2) \otimes (\vec{p}_g - \vec{x}_i), \\ \vec{x}_i \leftarrow \vec{x}_i + \vec{v}_i; \end{cases}$$

6: If a criterion is met (usually a sufficiently good fitness or a maximum number of iteration), exit loop;
7: **End loop**

In the algorithm, $\vec{U}(0,\phi_i)$ means a vector of random numbers uniformly distributed in $[0,\phi_i]$, $\vec{x}_i, \vec{v}_i$ means the current position and velocity respectively, and $\vec{p}_i, \vec{p}_g$ means the previous best position and the global best position.

In this study, the BPNN assisted PSO is used to obtain the optimal arrangement of holes and the flowchart is shown in Fig. 2 where an adaptive subdomain partition strategy is proposed to decide the number of holes should be used. Moreover, the BPNN is used to construct the model of fitness function value, so that the fitness function value can be obtained efficiently. The more details are shown in the following sections.

### 3.3.1 Fitness function

The PSO is used to find the optimal design variables (the size, position and number of holes) to make the real crack path match with the specified path. Assume an edge crack in a plate as shown in the left of Fig. 3, and then a specified crack path is given as shown in Fig. 6.

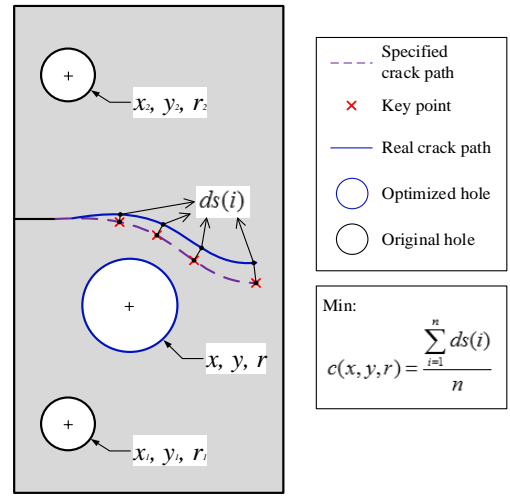

Fig. 6 The description of the optimization

Obviously, the specified crack path is defined by a set of key points and the ideal situation is to control the real crack cross through these points. Although it is difficult to be achieved, the optimal path might be found by optimization method. Thus, this optimization problem is formulated as

*Minimize:*

$$c(x,y,r) = \frac{\sum_{i=1}^{n} ds(i)}{n},  \qquad (7)$$

*Subject to:*

$$\mathbf{KU} = \mathbf{F}, \qquad (8)$$

$$ds(i) = f(x,y,r), \qquad (9)$$

$$\sqrt{(x-x_1)^2 + (y-y_1)^2} > r + r_1, \qquad (10)$$

$$\sqrt{(x-x_2)^2 + (y-y_2)^2} > r + r_2, \qquad (11)$$

$$x_{min} + r < x < x_{max} - r, \qquad (12)$$

$$y_{min} + r < y < y_{max} - r, \qquad (13)$$

where $c(x,y,r)$ is the fitness function and the purpose is to minimize the fitness function. Moreover, $ds(i)$ means the minimum distance from the key point $i$ to the real crack path, and $x_{min}, x_{max}, y_{min}, y_{max}$ are used to define the design space. Moreover, the crack propagation path should be calculated by the reanalysis based X-FEM as shown in Fig. 2, so the optimization must be subjected to the equilibrium equation and the $ds(i)$ is determined by the size, position or number of holes. It is obvious that the fitness function means the average distance from the key points to the real crack path, and it can be used to define the fitting accuracy between real crack and specified paths.

### 3.3.2 Adaptive subdomain partition strategy for design space

The adaptive subdomain partition strategy is proposed to determine the number of holes should be used. The main idea of the adaptive subdomain partition strategy is to divide the design space into some subdomains, and generate a hole in each subdomain. Generally, the less holes used the more convenient for processing, so the number of subdomains will be increased gradually. In this study, the number of subdomains will be started from one to two, three, four and so on. The process of the adaptive subdomain partition strategy is demonstrated in Fig. 7. First, a threshold value $\varepsilon$ needs to be defined, which is the target value of fitness function $c(x,y,r)$. Then the threshold value $\varepsilon$ is used to guarantee the fitting accuracy is available for engineering problems. When the minimal value of objective function $c(x,y,r) > \varepsilon$, it means the design space needs to be subdivided further, otherwise, the optimal solution is obtained. The details of this strategy are described as following:

| The adaptive subdomain partition strategy |
|---|
| 1: Initialize the number of subdomains in design space $n$ by $n=1$; |
| 2: **Loop** |
| 3: Calculate the fitness function $c(x,y,r)$ by Eq.(7); |
| 4: Find the minimum value $c_{min}$ of fitness function; |
| 5: If a criterion is met, exit loop. If not, $n=n+1$; The criterion is $$c_{min} = \min(c(x,y,r)) < \varepsilon;$$ |
| 6: Divide the design space into $n$ subdomains; |
| 7: **End loop** |

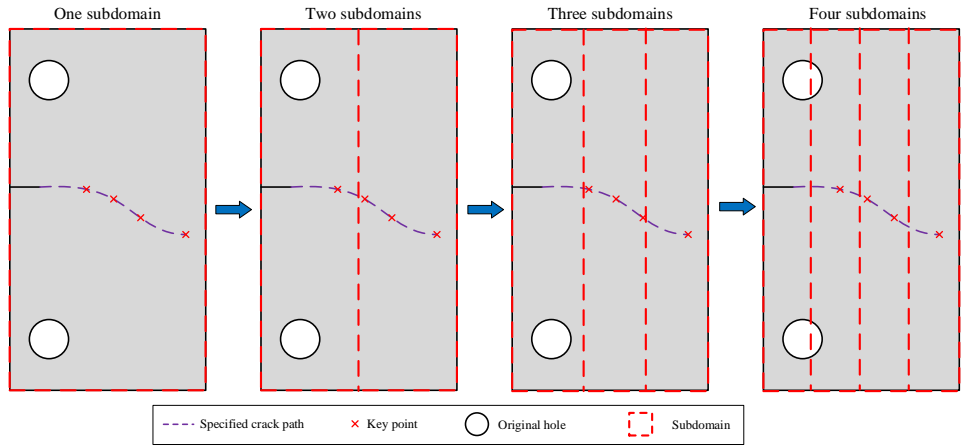

Fig. 7 An illustration of the adaptive subdomain partition strategy

### 3.3.3 Back propagation neural network

As mentioned above, the PSO needs to evaluate the fitness of each particle in every generation. Usually, the fitness could be evaluated by fitness function, such as Eq.(7), but it is time consumed, because the fitness needs to be evaluated by the X-FEM for each particle and the X-FEM is an iterative process. Therefore, we suggested a BPNN assisted PSO that the BPNN is used to construct the model of fitness function due to its flexible nonlinear modeling capability with strong adaptability (Ticknor, 2013). The BPNN is one of the popular artificial neural networks (L. Wang et al., 2006). It's a type of multi-layered feed-forward neural network that minimizes an error backward while information is transmitted forward (G. Zhang et al., 1998)and only one single hidden BPNN layer is enough to approximate any nonlinear function with arbitrary precision (Aslanargun et al., 2007). Therefore, the BPNN has been widely used in many fields and many intelligent evolution algorithms have also been used to select the initial connection weights and thresholds of BPNN, such as genetic algorithm (GA) (Irani et al., 2011), PSO (J.-R. Zhang et al., 2007) and so on.

In this study, the BPNN is used to forecast the fitness function value of PSO, so that the optimal solution can be found more efficiently than the popular PSO. Generally, a BPNN consists of an input layer, an output layer and several hidden layers as shown in Fig. 8. A systematic theory can be found in the literature (G. Zhang et al., 1998; L. Zhang et al., 2002). Furthermore, in order to forecast the fitness function value, a set of sample data should be used to construct a BPNN, then the BPNN need to be trained, and finally the trained BPNN can be used to forecast the fitness function value, the flowchart of BPNN assisted PSO is shown in Fig. 9.

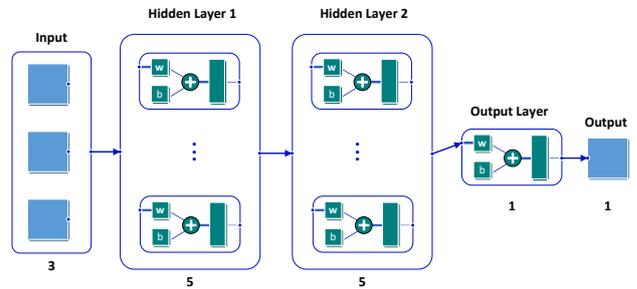

Fig. 8 Double hidden layers BPNN structure

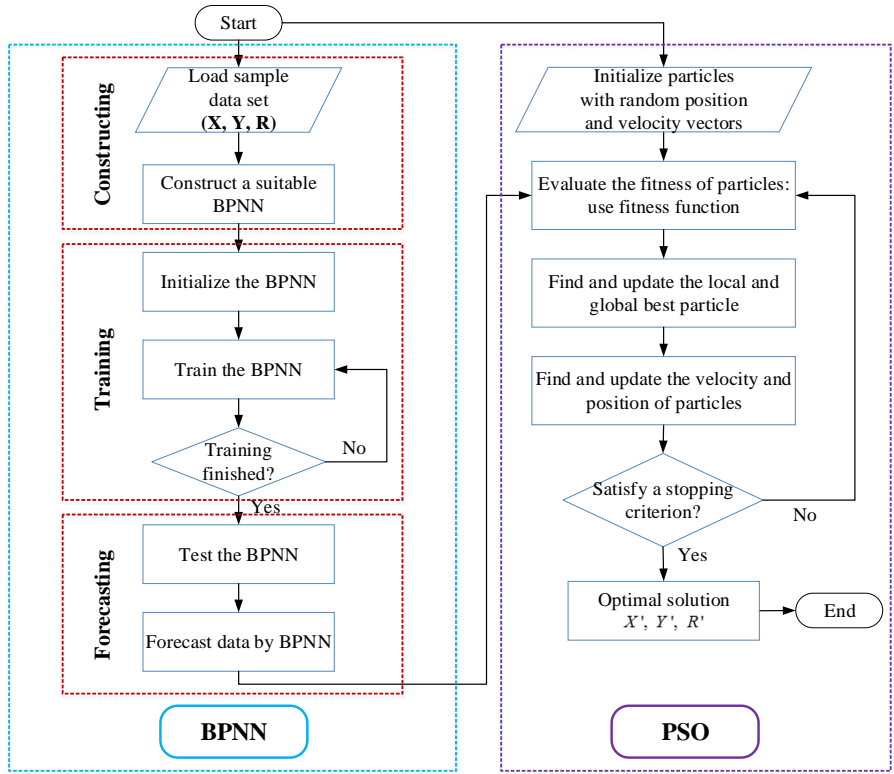

Fig. 9 Flowchart of BPNN assisted PSO

## 3.4 The accuracy and efficiency of reanalysis based X-FEM

Consider that the optimization need a large sample points and the calculation of the crack propagation path is an iterative process, the computational cost is commonly expensive. Therefore, an efficient reanalysis based X-FEM is used to calculate the crack propagation path, in which a reanalysis solver is used to solve the equilibrium equations efficiently. The reanalysis based X-FEM is used to predict the response of the current iteration by using the information of the first iteration. It avoids the full analysis after the first iteration, and the response of the subsequent iteration can be efficiently obtained.

In this study, an exact reanalysis named decomposed updating reanalysis (DUR) method is used to solve the equilibrium equations, the briefly introduction of this method is as following, more details can be found from the reference (Cheng et al., 2017).

Assume that the equilibrium equation of the X-FEM is

$$\mathbf{K}^{(i)}\mathbf{U}^{(i)} = \mathbf{F}^{(i)}, \tag{14}$$

where $\mathbf{U}^{(i)}$ is the displacement of the $i$-th iteration, and the equilibrium equation of the first iteration can be given as

$$\mathbf{K}^{(1)}\mathbf{U}^{(1)} = \mathbf{F}^{(1)}. \tag{15}$$

Define that

$$\begin{aligned} \mathbf{U}^{(i)} &= \mathbf{U}^{(1)} + \Delta\mathbf{U} \quad (i>1) \\ \boldsymbol{\delta} &= \mathbf{F}^{(i)} - \mathbf{K}^{(i)}\mathbf{U}^{(1)} \quad (i>1) \end{aligned}, \tag{16}$$

then substitute Eq.(16) into Eq.(14), obtained

$$\mathbf{K}^{(i)}\Delta\mathbf{U} = \boldsymbol{\delta}. \tag{17}$$

Consider that the increment of crack is very small in each iteration, so only small part of $\boldsymbol{\delta}$ should be nonzero. Based on this property, divide the $\mathbf{U}^{(i)}$ into two parts: unbalanced and balanced equations, according to Eq.(18):

$$\boldsymbol{\Delta} = sum\left(\left|\mathbf{K}^{(i)} - \mathbf{K}^{(1)}\right|\right) + \left|\boldsymbol{\delta}\right|. \tag{18}$$

If $\left|\boldsymbol{\Delta}(j)\right| > 0$, the $j$-th DOF is unbalanced, otherwise the $j$-th DOF is balanced.

Therefore, the Eq.(17) can be rewritten as

$$\begin{bmatrix} \mathbf{K}^{(i)}_{mm} & \mathbf{K}^{(i)}_{mn} \\ \mathbf{K}^{(i)}_{nm} & \mathbf{K}^{(i)}_{nn} \end{bmatrix} \begin{Bmatrix} \Delta\mathbf{U}_m \\ \Delta\mathbf{U}_n \end{Bmatrix} = \begin{Bmatrix} \mathbf{0} \\ \boldsymbol{\delta}_n \end{Bmatrix}, \tag{19}$$

where $m$ is the number of balanced DOFs, and $n$ is the number of unbalanced DOFs.

Equation (19) can be rewritten as

$$\mathbf{K}^{(i)}_{mm}\Delta\mathbf{U}_m + \mathbf{K}^{(i)}_{mn}\Delta\mathbf{U}_n = \mathbf{0} \tag{20}$$

and

$$\mathbf{K}^{(i)}_{nm}\Delta\mathbf{U}_m + \mathbf{K}^{(i)}_{nn}\Delta\mathbf{U}_n = \boldsymbol{\delta}_n. \tag{21}$$

Obviously, Eq.(20) is a homogeneous equation set which has infinite solutions, and the solution $\mathbf{U}^{(1)}$ is one of them. Calculate the fundamental solution system of Eq. (20) by

$$\mathbf{B} = \begin{bmatrix} -\mathbf{K}^{(i)\,-1}_{mm}\mathbf{K}^{(i)}_{mn} \\ \mathbf{E}_{nn} \end{bmatrix}, \tag{22}$$

where $\mathbf{E}_{nn}$ is a rank-$n$ unit matrix.

Define the general solution of Eq.(20) is
$$\Delta\mathbf{U} = \mathbf{B}\mathbf{y}, \tag{23}$$
where $\mathbf{y}$ is a dimension-$n$ vector. Then, substitute Eq.(23) into Eq.(21) to find a unique solution of Eq.(17), obtain

$$\left(\mathbf{K}^{(i)}_{nn} - \mathbf{K}^{(i)}_{nm}\mathbf{K}^{(i)\,-1}_{mm}\mathbf{K}^{(i)}_{mn}\right)\mathbf{y} = \boldsymbol{\delta}_n. \tag{24}$$

Solve Eq.(24), $\mathbf{y}$ can be obtained, and then $\Delta\mathbf{U}$ can be obtained by Eq.(23). Sequentially, the $\mathbf{U}^{(i)}$ can be obtained by Eq.(16).

In order to test the accuracy and efficiency of the reanalysis based X-FEM, an edge crack in a plate with a hole which mentioned in section 3.2 has been calculated by the reanalysis based X-FEM. The comparison between reanalysis, full analysis and experimental results (Giner et al., 2009) as shown in Fig. 10 and Fig. 11. Moreover, the average errors and computational cost are listed in Tab. 1. It can be found that the reanalysis based X-FEM is accurate and it saves about 13 times computational cost than the full analysis. Therefore, the reanalysis based X-FEM is an accurate method for crack quasi-static propagation problems with high efficiency.

Tab. 1 Performance comparison of edge crack in a plate with a hole by DUR and full analysis

| Computational Time/$s$ | | Average Errors | |
| --- | --- | --- | --- |
| DUR | Full analysis | Displacement | Von Mises Stress |
| 2.047 | 26.781 | 4.8916e-13 | 1.7356e-12 |

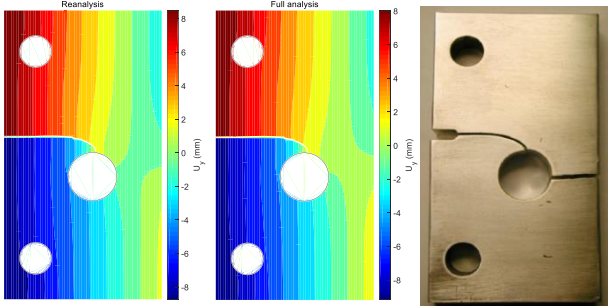

Fig. 10 The displacement comparison between reanalysis, full analysis and experimental results

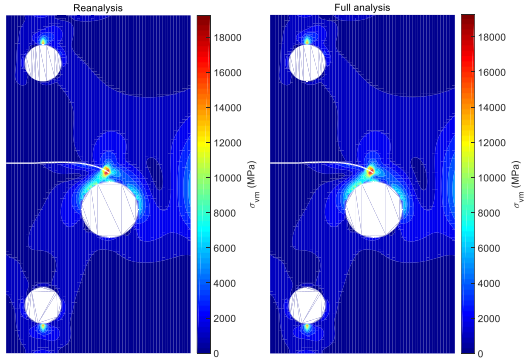

Fig. 11 The stress comparison between reanalysis and full analysis

## 4 Numerical examples

In order to test feasibility of the CCP method, two examples are tested by the proposed methods. These two cases involve a simple case and an engineering case.

### 4.1 An edge crack optimization in a plate

As shown in the left figure of Fig. 12, an edge crack plate is considered where the initial crack length $a_0 = 10mm$, and the uniformed load $q = 200N/mm$. The parameters of material are given as following: the Young's modulus $E_1 = 7.17 \times 10^4 MPa$, the Poisson's ratio $v = 0.33$, and a plane strain state is considered. Assume the increment of propagation $\Delta a = 1mm$, then the crack will propagate along a straight line like the middle figure of Fig. 12 if there with no holes. However, if we want to make the crack propagate along the specified path as shown in the right figure of Fig. 12, how could we realize it?

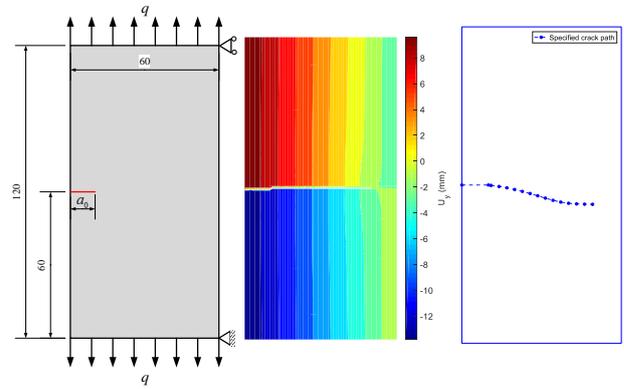

Fig. 12 An edge crack optimization in a plate

As mentioned above, some holes will be added to control the crack path, and the CCP method should be used to find the suitable number, position and size of these holes. As shown in Fig. 9, we need obtain a sample data set firstly. Then a BPNN should be constructed by the training sample with 1000 groups of sample data. After that, the BPNN will be trained and finally the testing sample with 500 groups of sample data should be used to test the performance of BPNN. Here a double hidden layers BPNN is used and each hidden layer has 5 nodes. The regression of BPNN is shown in Fig. 13, and the left is the regression of training sample while the right is the regression of testing sample. It can be found that the performance of BPNN is acceptable.

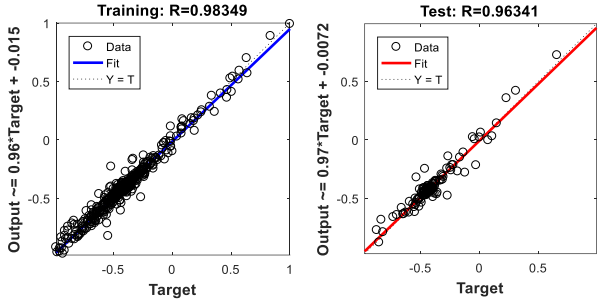

Fig. 13 The regression of BPNN

After the BPNN has been trained, the optimal solution should be obtained by the BPNN assisted PSO and popular PSO with 40 particles. The optimization procedure of each method is presented in Fig. 14. It can be found that the BPNN assisted PSO reaches convergence more quickly than the popular PSO while the popular PSO can obtain a smaller fitness function value. However, the BPNN assisted PSO can save much more computational time than the popular PSO because the fitness function value can be obtained by the BPNN rather than X-FEM method. The optimal solutions of them are listed in Tab. 2. Moreover, the results of crack propagation path are shown in Fig. 15 and Fig. 16 where the key points are used to define the specified crack path. The results indicate that the crack did propagate along the specified path.

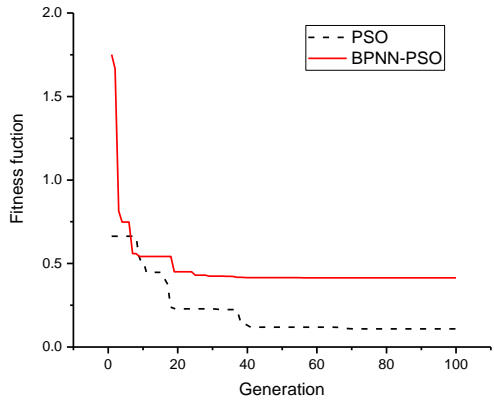

Fig. 14 Comparison between optimization procedures of BPNN-PSO and popular PSO

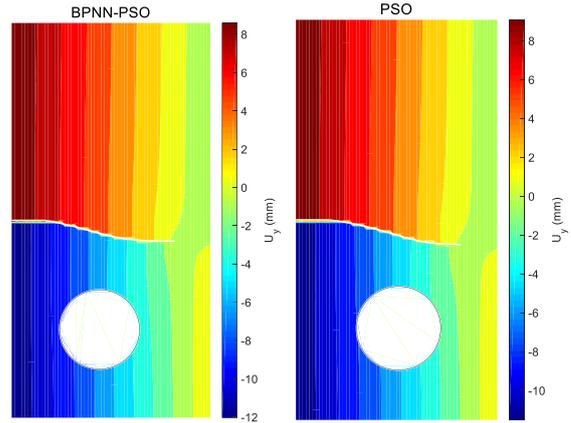

Fig. 15 The displacement results of optimal solution

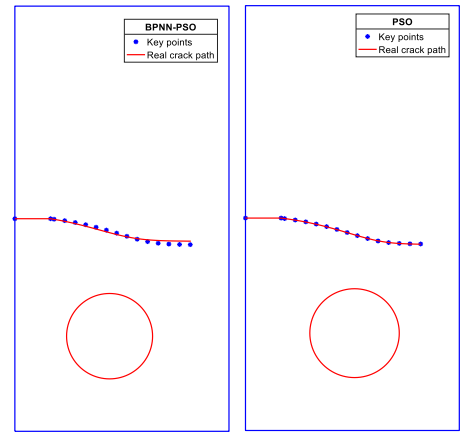

Fig. 16 Comparison of specified crack path and real crack path

### 4.2 The bottom plate of bearing pedestal

As mentioned before, the purpose of this study is to control the crack keep away from the critical domain of engineering structure to avoid failure. In this case, a bottom plate of bearing pedestal is considered as shown in Fig. 17 and all the parameters of material are the same as section 4.1. Assume a crack is in the edge of bottom plate as shown in Fig. 18, where the initial crack length $a_0 = 5mm$, and the uniformed load $q = 20 N/mm$.

Then the crack will propagate and cross the threaded hole like the right figure of Fig. 18, and this will lead the failure of bearing pedestal. Therefore, we need to control the crack propagation path to make it not cross the threaded holes just like Fig. 19, where the left figure is path A while the right is path B.

Tab. 2 Optimal solutions of BPNN assisted PSO and popular PSO

| Optimization | Design variables (*mm*) | | | Fitness function |
|---|---|---|---|---|
| | X | Y | R | |
| BPNN-PSO | 26.5407 | 26.8463 | 12.0507 | 0.4144 |
| PSO | 30.6038 | 27.4947 | 12.5502 | 0.1084 |

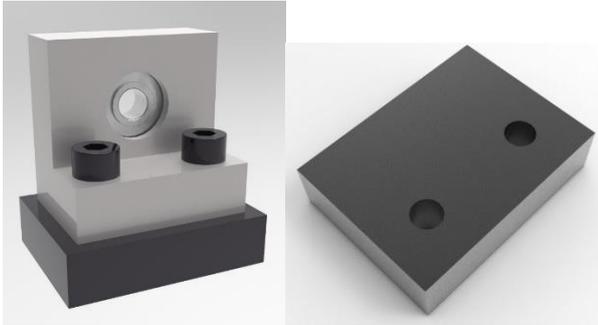

Fig. 17 A bottom plate of bearing pedestal

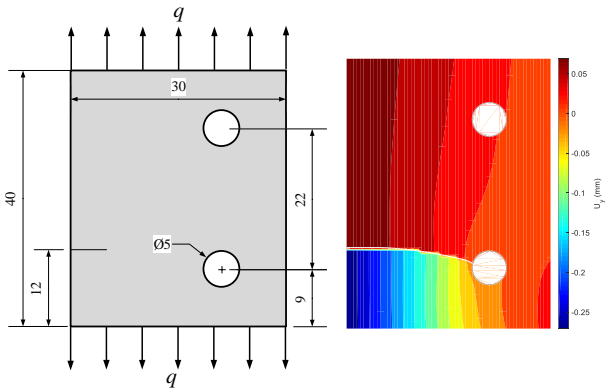

Fig. 18 An edge crack in the bottom plate

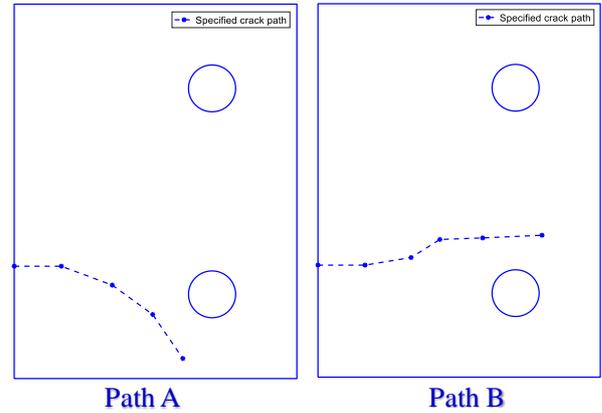

Fig. 19 The specified crack paths of bottom plate

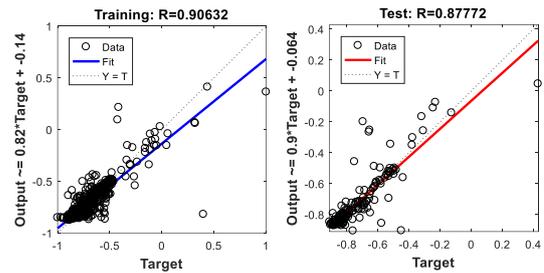

Fig. 20 The regression of BPNN

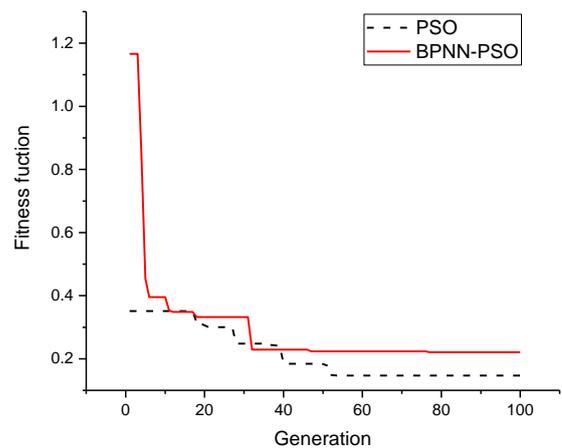

Fig. 21 Comparison between optimization procedures of BPNN-PSO and popular PSO of path A

For the path A, a BPNN has been constructed and trained by the training samples with 1000 groups of data and 500 groups of testing data have been tested the performance of BPNN. A double hidden layers BPNN is also used. The regression of BPNN is shown in Fig. 20. Then the optimal solution should be found by BPNN assisted PSO and popular PSO with 40 particles, the optimization procedure is shown in Fig. 21 and the optimal solutions are listed in Tab. 3. Moreover, the results of crack propagation path are shown in Fig. 22 and Fig. 23 where the key points are used to define the specified crack path. The results indicate that the crack did propagate along the specified path.

Tab. 3 Optimal solutions of BPNN assisted PSO and popular PSO

| Optimization | Design variables (*mm*) | | | Fitness function |
|---|---|---|---|---|
| | X | Y | R | |
| BPNN-PSO | 7.1572 | 3.1728 | 2.7756 | 0.2231 |
| PSO | 8.3984 | 3.9024 | 3.0252 | 0.1453 |

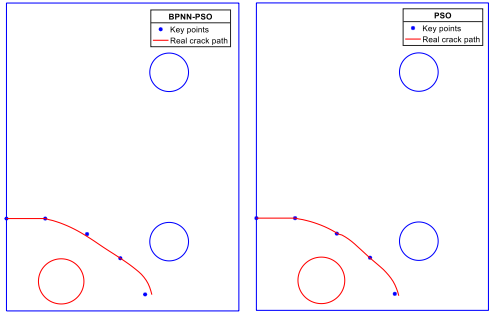

Fig. 22 Comparison of specified crack path and real crack path

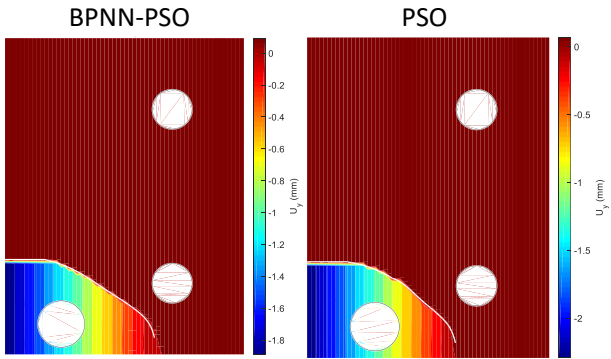

Fig. 23 The displacement results of optimal solution

As for the path B, the same operations have been carried out, and the optimal result by using only one hole is shown in Fig. 24. It can be found that the solution is not very nice by only one hole, so the adaptive subdomain partition strategy divides the design space into two sub-spaces, and for two holes, the optimal solution is much better. For two holes optimization, the regression of BPNN is shown in Fig. 25, where the size of training sample is 1000 and the size of testing sample is 500. It should be noted that the samples in center domain is not satisfied with the constraints Eqs. (10-13).Therefore, there are no training and testing samples. The optimization procedure is shown in Fig. 26 and the optimal solutions are listed in Tab. 4. Moreover, the results of optimal solution are shown in Fig. 27 and Fig. 28. The results indicate that the crack did propagate along the specified path.

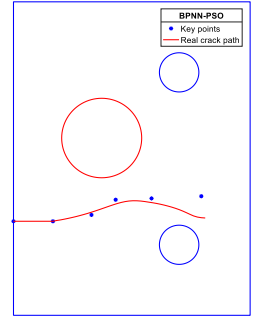

Fig. 24 The optimal result for only one hole

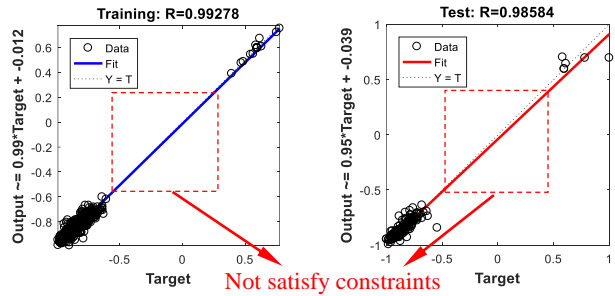

Fig. 25 The regression of BPNN

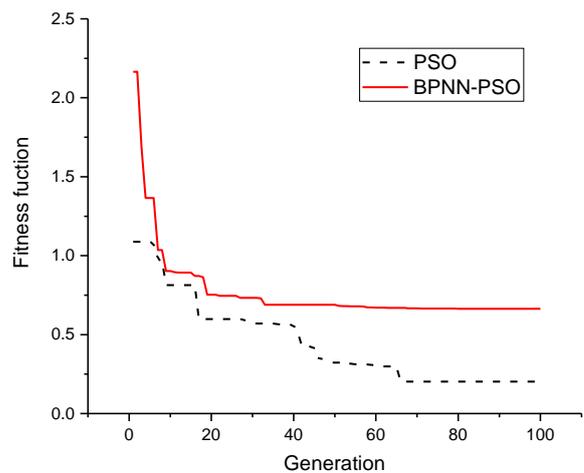

Fig. 26 Comparison between optimization procedures of BPNN-PSO and popular PSO of path B

Tab. 4 Optimal solutions of BPNN assisted PSO and popular PSO

| Optimization | Design variables (*mm*) | | | | | | Fitness function |
| --- | --- | --- | --- | --- | --- | --- | --- |
| | $X_1$ | $Y_1$ | $R_1$ | $X_2$ | $Y_2$ | $R_2$ | |
| BPNN-PSO | 10.1674 | 20.3498 | 4.5053 | 24.1869 | 21.2344 | 3.8063 | 0.66365 |
| PSO | 10.3278 | 20.4236 | 4.5921 | 24.7407 | 21.2991 | 3.6068 | 0.20189 |

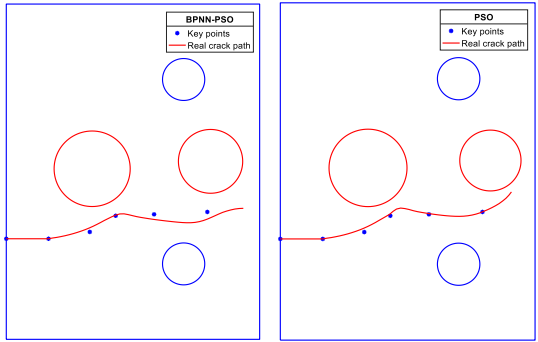

Fig. 27 Comparison of specified crack path and real crack path

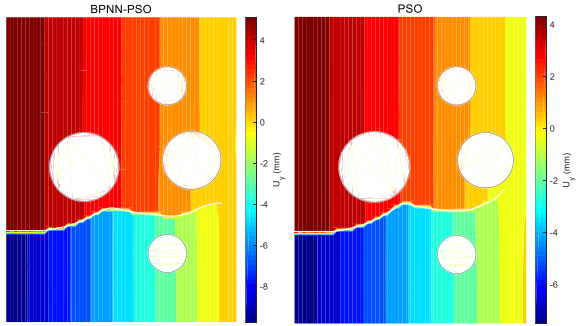

Fig. 28 The displacement results of optimal solution

### 4.3 Analysis of computational cost

Two numerical examples have been tested in this section by both the BPNN assisted PSO and popular PSO methods. In order to compare the performance of BPNN assisted and popular PSOs, the CPU running time has been recorded and all simulations were performed on an Intel(R) Core(TM) i7-5820K 3.30GHz CPU with 32GB of memory within MATLAB R2016b in x64 Windows 7. Firstly, a comparison of computational scale between the BPNN assisted PSO and popular PSO is shown in Tab. 5, where the computational data means the data set which needs to be calculated during the optimization process and it can be calculated by accumulating all particles in all generations until the optimization converges. The sample data is used to construct the BPNN model, so there is no sample data for the popular PSO. For a sample point, the computational time cost in solving the equilibrium equations is listed in Tab. 6. It can be found that the reanalysis solver DUR is more efficient than the full analysis. Moreover, the modeling, optimization and total time cost by the DUR and full analysis are listed in Tab. 7 and Tab. 8 respectively, where the modeling time is the cost of constructing BPNN model and the optimization time is the cost of optimization process. The comparison between the DUR and full analysis is shown in Fig. 29.

From the above results, it can be found that the accuracy of BPNN determine the performance of the BPNN assisted PSO. However, in term of efficiency, it prevails. Therefore, the tradeoff between efficiency and accuracy is important for selection. Moreover, the reanalysis solver DUR saves much computational cost in solving the equilibrium equations. Furthermore, the comparison of specified crack path and real crack path indicates that the crack did propagate along the specified path, so the CCP method is available to control the crack propagation path.

Tab. 5 Comparison of computational scale between the BPNN assisted PSO and popular PSO

| | Convergent generation | | Size of sample data | | Size of computational data | |
| --- | --- | --- | --- | --- | --- | --- |
| | BPNN-PSO | PSO | BPNN-PSO | PSO | BPNN-PSO | PSO |
| Case 1 | 20 | 40 | 1000 | -- | 800 | 1600 |
| Case 2-1 | 31 | 50 | 1000 | -- | 1240 | 2000 |
| Case 2-2 | 32 | 67 | 1000 | -- | 1280 | 2680 |

Tab. 6 Comparison of computational time cost between DUR and full analysis for on data point

|  | Computational time/*s* | |
|---|---|---|
|  | DUR | Full analysis |
| Case 1 | 3.062 | 37.796 |
| Case 2-1 | 1.984 | 20.578 |
| Case 2-2 | 1.863 | 19.965 |

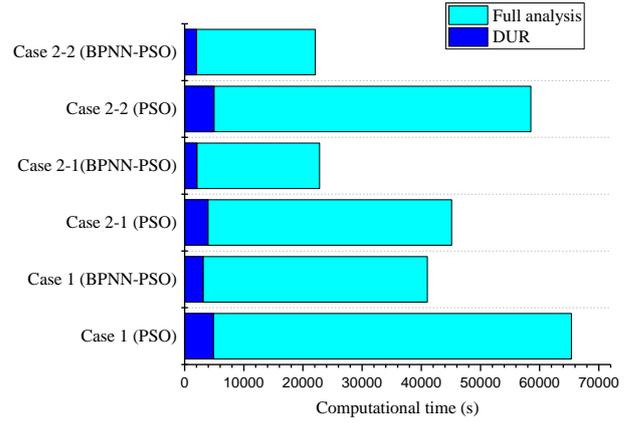

Fig. 29 Comparison of computational time cost between DUR and full analysis

Tab. 7 Computational time of the BPNN-PSO and PSO by reanalysis solver DUR

|  | Modeling time/*s* | | Optimization time/*s* | | Total time/*s* | |
|---|---|---|---|---|---|---|
|  | BPNN-PSO | PSO | BPNN-PSO | PSO | BPNN-PSO | PSO |
| Case 1 | 3062 | -- | 76 | 4899 | 3138 | 4899 |
| Case 2-1 | 1984 | -- | 121 | 3968 | 2105 | 3968 |
| Case 2-2 | 1863 | -- | 129 | 4992 | 1992 | 4992 |

Tab. 8 Computational time of the BPNN-PSO and PSO by reanalysis solver full analysis

|  | Modeling time/*s* | | Optimization time/*s* | | Total time/*s* | |
|---|---|---|---|---|---|---|
|  | BPNN-PSO | PSO | BPNN-PSO | PSO | BPNN-PSO | PSO |
| Case 1 | 37796 | -- | 81 | 60473 | 37877 | 60473 |
| Case 2-1 | 20578 | -- | 118 | 41156 | 20696 | 41156 |
| Case 2-2 | 19965 | -- | 132 | 53506 | 20097 | 53506 |

## 5 Conclusions

In this study, the controllable crack propagation (CCP) method is proposed to control the crack propagation path. Maybe the crack propagation is unavoidable, but the crack propagation path can be specified by the CCP method, so the critical domain of structure will not be crossed. The main idea of CCP is to control the crack propagation by arranging some holes in the design domain, so the crack propagation path can be driven by a suitable arrangement. The CCP method is a closed loop optimization method which integrating the BPNN, PSO, reanalysis based X-FEM, adaptive subdomain partition strategy and other techniques. Two optimization methods are used to find the suitable arrangement which including the number, position and size of holes. Firstly, a BPNN assisted PSO is suggested. In this method, the fitness function value of PSO can be forecasted by the BPNN rather than X-FEM, so efficiency of BPNN assisted PSO is much higher than popular PSO. Moreover, considering the optimization iteration is a time consuming process, an efficient reanalysis based X-FEM is used to calculate the crack propagation path, in which reanalysis methods are used to solve the equilibrium equations efficiently. Furthermore, in order to improve the fitting accuracy between real crack path and specified crack path, the adaptive subdomain partition strategy is suggested to decide the number of holes should be used. Numerical examples indicate that the CCP method can control the crack propagation along the specified path well, and the fitting accuracy between real crack and specified paths is available.

## Acknowledgements

This work has been supported by Project of the National Key R&D Program of China 2017YFB0203701 and the National Natural Science Foundation of China under the Grant Numbers 11572120.

## References

Abdelaziz, Y., & Hamouine, A. (2008). A survey of the extended finite element. *Computers & Structures, 86*(11-12), 1141-1151.

Ahmed, A., van der Meer, F. P., & Sluys, L. J. (2012). A geometrically nonlinear discontinuous solid-like shell element (DSLS) for thin shell structures. *Computer Methods in Applied Mechanics and Engineering, 201–204*, 191-207.

Amini, F., Hazaveh, N. K., & Rad, A. A. (2013). Wavelet PSO-Based LQR Algorithm for Optimal Structural Control Using Active Tuned Mass Dampers. *Computer-Aided Civil and Infrastructure Engineering, 28*(7), 542-557.

Amiri, G. G., Rad, A. A., Aghajari, S., & Hazaveh, N. K. (2012). Generation of Near-Field Artificial Ground Motions Compatible with Median-Predicted Spectra Using PSO-Based Neural Network and Wavelet Analysis. *Computer-Aided Civil and Infrastructure Engineering, 27*(9), 711–730.

Areias, P., & Belytschko, T. (2005). Non-linear analysis of shells with arbitrary evolving cracks using XFEM. *International Journal for Numerical Methods in Engineering, 62*(3), 384-415.

Aslanargun, A., Mammadov, M., Yazici, B., & Yolacan, S. (2007). Comparison of ARIMA, neural networks and hybrid models in time series: tourist arrival forecasting. *Journal of Statistical Computation and Simulation, 77*(1), 29-53.

Battaïa, O., & Dolgui, A. (2013). A taxonomy of line balancing problems and their solutionapproaches. *International Journal of Production Economics, 142*(2), 259-277.

Belytschko, T., & Black, T. (1999). Elastic crack growth in finite elements with minimal remeshing. *International Journal for Numerical Methods in Engineering, 45*(5), 601-620.

Belytschko, T., Chen, H., Xu, J., & Zi, G. (2003). Dynamic crack propagation based on loss of hyperbolicity and a new discontinuous enrichment. *International Journal for Numerical Methods in Engineering, 58*(12), 1873-1905.

Belytschko, T., Gracie, R., & Ventura, G. (2009). A review of extended/generalized finite element methods for material modeling. *Modelling and Simulation in Materials Science and Engineering, 17*(4), 043001.

Belytschko, T., Moës, N., Usui, S., & Parimi, C. (2001). Arbitrary discontinuities in finite elements. *International Journal for Numerical Methods in Engineering, 50*(4), 993-1013.

Bouchard, P. O., Bay, F., & Chastel, Y. (2003). Numerical modelling of crack propagation: automatic remeshing and comparison of different criteria. *Computer Methods in Applied Mechanics and Engineering, 192*(35-36), 3887-3908.

Branco, R., Antunes, F. V., & Costa, J. D. (2015). A review on 3D-FE adaptive remeshing techniques for crack growth modelling. *Engineering Fracture Mechanics, 141*, 170-195.

Chatterjee, S., Sarkar, S., Hore, S., Dey, N., Ashour, A. S., & Balas, V. E. (2016). Particle swarm optimization trained neural network for structural failure prediction of multistoried RC buildings. *Neural Computing and Applications, 28*(8), 2005-2016.

Cheng, Z., & Wang, H. (2017). A fast and exact computational method for crack propagation based on extended finite element method. *arXiv preprint arXiv:1708.01610*.

Chessa, J., Smolinski, P., & Belytschko, T. (2002). The extended finite element method (XFEM) for solidification problems. *International Journal for Numerical Methods in Engineering, 53*(8), 1959-1977.

Delice, Y., Kızılkaya Aydoğan, E., Özcan, U., & İlkay, M. S. (2014). A modified particle swarm optimization algorithm to mixed-model two-sided assembly line balancing. *Journal of Intelligent Manufacturing, 28*(1), 23-36.

Dolbow, J., & Belytschko, T. (1999). A finite element method for crack growth without remeshing. *International Journal for Numerical Methods in Engineering, 46*(1), 131-150.

Eberhart, & Shi, Y. (2001). *Particle swarm optimization: developments, applications and resources.* Paper presented at the Evolutionary Computation, 2001. Proceedings of the 2001 Congress on.

Erdogan, F., & Sih, G. (1963). On the crack extension in plates under plane loading and transverse shear. *Journal of basic engineering, 85*(4), 519-527.

Fries, T. P., & Belytschko, T. (2010). The extended/generalized finite element method: an overview of the method and its applications. *International Journal for Numerical Methods in Engineering, 84*(3), 253-304.

Giner, E., Sukumar, N., Tarancón, J. E., & Fuenmayor, F. J. (2009). An Abaqus implementation of the extended finite element method. *Engineering Fracture Mechanics, 76*(3), 347-368.

Gravouil, A., Moës, N., & Belytschko, T. (2002). Non-planar 3D crack growth by the extended finite element and level sets—Part II: Level set update. *International Journal for Numerical Methods in Engineering, 53*(11), 2569-2586.

Gu, Y. T., Wang, W., Zhang, L. C., & Feng, X. Q. (2011). An enriched radial point interpolation method (e-RPIM) for analysis of crack tip fields. *Engineering Fracture Mechanics, 78*(1), 175-190.

He, G., Wang, H., Li, E., Huang, G., & Li, G. (2015). A multiple-GPU based parallel independent coefficient reanalysis method and applications for vehicle design. *Advances in Engineering Software, 85*, 108-124.

Huang, G., Wang, H., & Li, G. (2013). A reanalysis method for local modification and the application in large-scale problems. *Structural and Multidisciplinary Optimization, 49*(6), 915-930.

Huynh, D., & Belytschko, T. (2009). The extended finite element method for fracture in composite materials. *International Journal for Numerical Methods in Engineering, 77*(2), 214-239.

Irani, R., & Nasimi, R. (2011). Evolving neural network using real coded genetic algorithm for permeability estimation of the reservoir. *Expert Systems with Applications, 38*(8), 9862-9866.

Kennedy, J., & Eberhart, R. (1995). *Particle swarm optimization.* Paper presented at the IEEE International Conference on Neural Networks, 1995. Proceedings.

Kennedy, J., & Eberhart, R. C. (1997). *A discrete binary version of the particle swarm algorithm.* Paper presented at the IEEE International Conference on Systems, Man, and Cybernetics, 1997. Computational Cybernetics and Simulation.

Kirsch, U. (2002). *Design-Oriented Analysis of Structures*. Dordrecht: Kluwer Academic Publishers.

Liu, G. R., Nourbakhshnia, N., & Zhang, Y. W. (2011). A novel singular ES-FEM method for simulating singular stress fields near the crack tips for linear fracture problems. *Engineering Fracture Mechanics, 78*(6), 863-876.

Liu, H. F., Wu, B. S., & Li, Z. G. (2014). Method of Updating the Cholesky Factorization for Structural Reanalysis with Added Degrees of Freedom. *Journal of Engineering Mechanics, 140*(2), 384-392.

Ma, W., Wang, M., & Zhu, X. (2015). Hybrid particle swarm optimization and differential evolution algorithm for bi-level programming problem and its application to pricing and lot-sizing decisions. *Journal of Intelligent Manufacturing, 26*(3), 471-483.

Nguyen-Xuan, H., Liu, G. R., Bordas, S., Natarajan, S., & Rabczuk, T. (2013). An adaptive singular ES-FEM for mechanics problems with singular field of arbitrary order. *Computer Methods in Applied Mechanics and Engineering, 253*, 252-273.

Poli, R., Kennedy, J., & Blackwell, T. (2007). Particle swarm optimization. *Swarm Intelligence, 1*(1), 33-57.

Shabbir, F., & Omenzetter, P. (2015). Particle Swarm Optimization with Sequential Niche Technique for Dynamic Finite Element Model Updating. *Computer-Aided Civil and Infrastructure Engineering, 30*(5), 359-375.

Shi, Y., & Eberhart, R. (1998). *A modified particle swarm optimizer*: Springer Berlin Heidelberg.

Song, J.-H., Areias, P. M. A., & Belytschko, T. (2006). A method for dynamic crack and shear band propagation with phantom nodes. *International Journal* for *Numerical Methods in Engineering, 67*(6), 868-893.

Song, Q., Chen, P., & Sun, S. (2014). An exact reanalysis algorithm for local non-topological high-rank structural modifications in finite element analysis. *Computers & Structures, 143*, 60-72.

Stolarska, M., Chopp, D., Moës, N., & Belytschko, T. (2001). Modelling crack growth by level sets in the extended finite element method. *International Journal for Numerical Methods in Engineering, 51*(8), 943-960.

Sukumar, N., Chopp, D. L., Moës, N., & Belytschko, T. (2001). Modeling holes and inclusions by level sets in the extended finite-element method. *Computer Methods in Applied Mechanics and Engineering, 190*(46), 6183-6200.

Sukumar, N., Moës, N., Moran, B., & Belytschko, T. (2000). Extended finite element method for three-dimensional crack modelling. *International Journal for Numerical Methods in Engineering, 48*(11), 1549-1570.

Sun, R., Liu, D., Xu, T., Zhang, H., & Zuo, W. (2014). New Adaptive Technique of Kirsch Method for Structural Reanalysis. *AIAA Journal, 52*(3), 486-495.

Sun, W., & Xu, Y. (2016). Using a back propagation neural network based on improved particle swarm optimization to study the influential factors of carbon dioxide emissions in Hebei Province, China. *Journal of Cleaner Production, 112*, 1282-1291.

Tanaka, S., Suzuki, H., Sadamoto, S., Imachi, M., & Bui, T. Q. (2015). Analysis of cracked shear deformable plates by an effective meshfree plate formulation. *Engineering Fracture Mechanics, 144*, 142-157.

Ticknor, J. L. (2013). A Bayesian regularized artificial neural network for stock market forecasting. *Expert Systems with Applications, 40*(14), 5501-5506.

Tyagi, S. K., Yang, K., Tyagi, A., & Dwivedi, S. N. (2011). Development of a fuzzy goal programming model for optimization of lead time and cost in an overlapped product development project using a Gaussian Adaptive Particle Swarm Optimization-based approach. *Engineering Applications of Artificial Intelligence, 24*(5), 866-879.

Vagelis, P., & Manolis, P. (2011). A Hybrid Particle Swarm - Gradient Algorithm for Global Structural Optimization. *Computer-Aided Civil and Infrastructure Engineering, 26*(1), 48-68.

Wang, H., Zeng, Y., Li, E., Huang, G., Gao, G., & Li, G. (2016). "Seen Is Solution" a CAD/CAE integrated parallel reanalysis design system. *Computer Methods in Applied Mechanics and Engineering, 299*, 187-214.

Wang, L., Zeng, Y., Zhang, J., Huang, W., & Bao, Y. (2006). The criticality of spare parts evaluating model using artificial neural network approach. *Computational Science–ICCS 2006*, 728-735.

Zeng, Q., Liu, Z., Xu, D., Wang, H., & Zhuang, Z. (2016). Modeling arbitrary crack propagation in coupled shell/solid structures with X-FEM. *International Journal for Numerical Methods in Engineering, 106*(12), 1018-1040.

Zhang, G., Patuwo, B. E., & Hu, M. Y. (1998). Forecasting with artificial neural networks:: The state of the art. *International journal of forecasting, 14*(1), 35-62.

Zhang, J.-R., Zhang, J., Lok, T.-M., & Lyu, M. R. (2007). A hybrid particle swarm optimization–back-propagation algorithm for feedforward neural network training. *Applied Mathematics and Computation, 185*(2), 1026-1037.

Zhang, L., & Subbarayan, G. (2002). An evaluation of back-propagation neural networks for the optimal design of structural systems: Part II. Numerical evaluation. *Computer Methods in Applied Mechanics and Engineering, 191*(25), 2887-2904.

Zhuang, Z., & Cheng, B. B. (2011). Equilibrium state of mode-I sub-interfacial crack growth in bi-materials. *International Journal of Fracture, 170*(1), 27-36.

Zilian, A., & Legay, A. (2008). The enriched space–time finite element method (EST) for simultaneous solution of fluid–structure interaction. *International Journal for Numerical Methods in Biomedical Engineering, 75*(3).

Zuo, W., Xu, T., Zhang, H., & Xu, T. (2011). Fast structural optimization with frequency constraints by genetic algorithm using adaptive eigenvalue reanalysis methods. *Structural and Multidisciplinary Optimization, 43*(6), 799-810.